\documentclass{ieeeaccess}
\usepackage{cite}
\usepackage{amsmath,amssymb,amsfonts}
\usepackage{algorithmic}
\usepackage{graphicx}
\usepackage{textcomp}
\usepackage{url} 

\usepackage{booktabs}
\usepackage{makecell}
\usepackage{rotating}
\def\Rot#1{ \multicolumn{1}{l}{\rlap{\rotatebox{45}{#1}~}}}
\usepackage{arydshln}

\usepackage{bm}
\makeatletter
\AtBeginDocument{\DeclareMathVersion{bold}
\SetSymbolFont{operators}{bold}{T1}{times}{b}{n}
\SetSymbolFont{NewLetters}{bold}{T1}{times}{b}{it}
\SetMathAlphabet{\mathrm}{bold}{T1}{times}{b}{n}
\SetMathAlphabet{\mathit}{bold}{T1}{times}{b}{it}
\SetMathAlphabet{\mathbf}{bold}{T1}{times}{b}{n}
\SetMathAlphabet{\mathtt}{bold}{OT1}{pcr}{b}{n}
\SetSymbolFont{symbols}{bold}{OMS}{cmsy}{b}{n}
\renewcommand\boldmath{\@nomath\boldmath\mathversion{bold}}}
\makeatother

\def\BibTeX{{\rm B\kern-.05em{\sc i\kern-.025em b}\kern-.08em
    T\kern-.1667em\lower.7ex\hbox{E}\kern-.125emX}}

\begin{document}
\history{Date of publication xxxx 00, 0000, date of current version xxxx 00, 0000.}
\doi{10.1109/ACCESS.2024.0429000}

\title{A Comparative Study of Task Adaptation Techniques of Large Language Models for Identifying Sustainable Development Goals}
\author{
\uppercase{Andrea Cadeddu}\authorrefmark{1}, 
\uppercase{Alessandro Chessa}\authorrefmark{1}, 
\uppercase{Vincenzo De Leo}\authorrefmark{1,2}, 
\uppercase{Gianni Fenu}\authorrefmark{2}, 
\uppercase{Enrico Motta}\authorrefmark{3}, 
\uppercase{Francesco Osborne}\authorrefmark{3}, 
\uppercase{Diego Reforgiato Recupero}\authorrefmark{2}, 
\uppercase{Angelo Salatino}\authorrefmark{3}, 
\uppercase{Luca Secchi}\authorrefmark{2}
}

\address[1]{Linkalab s.r.l., Cagliari, Italy}
\address[2]{Department of Mathematics and Computer Science, University of Cagliari, Italy}
\address[3]{Knowledge Media Institute, The Open University, London, United Kingdom}

\markboth
{Author \headeretal: Preparation of Papers for IEEE TRANSACTIONS and JOURNALS}
{Author \headeretal: Preparation of Papers for IEEE TRANSACTIONS and JOURNALS}

\corresp{Corresponding author: Vincenzo De Leo (e-mail: vincenzo.deleo@linkalab.it).}

\begin{abstract}
In 2012, the United Nations introduced 17 Sustainable Development Goals (SDGs) aimed at creating a more sustainable and improved future by 2030. However, tracking progress toward these goals is difficult because of the extensive scale and complexity of the data involved. Text classification models have become vital tools in this area, automating the analysis of vast amounts of text from a variety of sources. Additionally, large language models (LLMs) have recently proven indispensable for many natural language processing tasks, including text classification, thanks to their ability to recognize complex linguistic patterns and semantics. This study analyzes various proprietary and open-source LLMs for a single-label, multi-class text classification task focused on the SDGs. Then, it also evaluates the effectiveness of task adaptation techniques (i.e., in-context learning approaches), namely Zero-Shot and Few-Shot Learning, as well as Fine-Tuning within this domain. The results reveal that smaller models, when optimized through prompt engineering, can perform on par with larger models like OpenAI's GPT (Generative Pre-trained Transformer).
\end{abstract}

\begin{keywords}
Large Language Models, Sustainable Development Goals, Text Classification
\end{keywords}

\titlepgskip=-21pt

\maketitle

\section{Introduction}
\label{sec:introduction}
\PARstart{T}{he} SDGs are a set of 17 interconnected global goals designed as a ``\textit{blueprint to achieve a better and more sustainable future for all}''\footnote{\url{https://www.un.org/sustainabledevelopment/sustainable-development-goals/}}. Established during the United Nations Conference on Sustainable Development in Rio de Janeiro in 2012, these goals are to be achieved by 2030~\cite{2013150612-FINAL-SDSN-Indicator-Report1,2017Data-for-Development-Full-Report}. They form the core of the 2030 Agenda for Sustainable Development, which was unanimously adopted by all United Nations Member States.

Monitoring and evaluating progress towards the United Nations' SDGs is a significant challenge today due to the scale and complexity of the data involved~\cite{HIRAI2022109605, RePEc:gam:jsusta:v:15:y:2023:i:4:p:3203-:d:1063481, Mishra2023, LaFleur2023}\footnote{\url{https://unece.org/sites/default/files/2021-04/2012761_E_web.pdf}}. Manual analysis methods are no longer sufficient to keep up with the fast and massive data generation in today's interconnected world~\cite{Nilashi2023Critical}.
In this context, text classification models have become essential tools, enabling the analysis of large volumes of textual data from various sources, including reports, news articles, social media, and institutional documents~\cite{10.1007/978-3-031-21743-2_21, Matsui2022, Guisiano2021}. These models help automatically identify and categorize content related to each SDG, supporting real-time monitoring and assessing how successful efforts related to SDGs are in delivering results. Consequently, they allow organizations to respond swiftly to crises, pinpoint urgent issues, and allocate resources effectively to address pressing SDGs~\cite{https://doi.org/10.1002/csr.2202, DBLP:journals/corr/abs-2005-14569}.
Moreover, classification models provide a data-driven basis for decision-making, enhancing transparency and accountability in sustainability efforts. This alignment not only helps organizations contribute meaningfully to global sustainability targets but also improves their reputation, meets regulatory requirements, and can lead to long-term economic benefits through enhanced efficiency and innovation. 
As a result, private enterprises are also recognizing the benefits of using classification models to align their strategies with the SDGs\footnote{\url{https://clarity.ai/} provides an interesting platform dedicated to evaluating the sustainability of funds and companies.}\footnote{\url{https://www.ibm.com/sustainability} showcases IBM’s framework for sustainability solutions.}. This alignment not only aids in making a positive sustainability impact but also enhances a company’s reputation and competitive edge by demonstrating a commitment to social and environmental goals.

In recent years, LLMs have emerged as a powerful solution, achieving state-of-the-art performance in various Natural Language Processing (NLP) tasks~\cite{10.5555/3295222.3295349}. These models excel in capturing complex linguistic patterns and semantics, demonstrating outstanding capabilities in applications such as text classification, sentiment analysis, document categorization, and language understanding tasks~\cite{devlin-etal-2019-bert, NEURIPS2020_1457c0d6}. They have also been evolving rapidly, with new and more effective LLMs being developed every few months over the past two years. In this context, it is essential to evaluate the performance of the latest generation of LLMs in identifying SDGs from text.

In this paper, we analyze several proprietary and open-source LLMs for a single-label multi-class text classification task specific to SDGs by using different task adaptation techniques. In particular, we aimed to determine whether smaller models, which require fewer resources and employ prompt engineering-based optimization techniques, could achieve results comparable to those of larger, well-known models like OpenAI's GPT. For this purpose, we established a benchmark dataset extracted from a larger dataset developed by the open-source initiative Open SDG (OSDG)\footnote{\url{https://open-sdg.org/}}. 
We employed different techniques commonly applied when optimizing LLMs for classification tasks, including in-context learning (Zero-Shot Learning and Few-Shot Learning) and Fine-Tuning. Additionally, we evaluated the impact of quantization by testing some quantized versions of well-known models to assess their performance on this task.

The findings provide interesting insights into how the model performs. Firstly, the best overall approach is a fine-tuned version of LLaMa-2 13B, achieving an F1 score of 92.4\%, followed by the fine-tuned versions of GPT-3.5-turbo (91.4\%) and Flan-T5-XXL (90.7\%). These results confirm that LLMs can achieve excellent performance in classifying SGDs. Interestingly, the fine-tuned version of LLaMa-3 8B (86.7\%) did not perform as well, being outperformed by alternatives such as Mistral 7B (88.1\%) and BERT (87.2\%). 
Different models excel with varying numbers of examples in a Few-Shot Learning setting. Specifically, LLaMa-3\_8B-Instruct performs best with only 3 examples (86.0\%), GPT-4.0 excels with 5 examples (86.9\%), and Mixtral-8x7B-Instruct is superior with up to 7 examples (88.1\%). 
Furthermore, the analysis of different approaches for Few-Shot Learning revealed that semantic similarity tends to work best for this task. 
Overall, the fact that fine-tuned versions of leading open models such as LLaMa-2 13B, Flan-T5-xxl, and Mistral 7B produced results comparable to or better than the fine-tuned version of GPT-3.5-turbo and the Few-Shot Learning versions of GPT-4 indicates that top results can be achieved with relatively small models when fine-tuned on high-quality data. Furthermore, some quantized solutions, such as the fine-tuned versions of LLaMa-2 13B 4-bit (85.7\%) and Zephyr 7B 4-bit (86.4\%), achieved highly competitive results, demonstrating their value in low-resource settings.

In summary, the main contributions of this paper are the following:
\begin{itemize}
\item A new benchmark for classification tasks within the domain of SDGs;
\item The evaluation of eighteen LLMs using this benchmark, including very recent solutions such as LLaMa 3;
\item A thorough study of various task adaptation techniques for LLMs, including Zero-Shot Learning, Few-Shot Learning (utilizing three different sample selection methods: random selection (``RANDOM''), semantic similarity (``SEM. SIMILARITY''), and semantic similarity across different classes (``SS WITH DC'')), and Fine-Tuning.
\end{itemize}

This work has been carried out in collaboration with Ovum S.r.l., an Italian startup specializing in software development for Artificial Intelligence, Cloud Computing, and Big Data. In particular, Ovum S.r.l. is interested in developing a tool to efficiently and objectively interpret large volumes of textual data through the lens of SDGs. 

The remainder of this paper is structured as follows. In Section \ref{sec:relatedwork}, we discuss previous research to contextualize the significance of our study. Section \ref{sec:background} details the specific task that we have tackled. Specifically, Section \ref{sec:EU_Dataset} describes the benchmark dataset used, while Section \ref{sec:Used_models} outlines the characteristics of the LLMs employed in our experiments. 
Section \ref{sec:experimentalevaluation} specifies the implementation details of each experiment. The results are presented and discussed in Section \ref{sec:results}. Finally, Section \ref{sec:conclusions} concludes our study and provides recommendations for researchers aiming to develop solutions that optimize the balance between performance and resource efficiency.

\section{Related Work}
\label{sec:relatedwork}

In this section, we explore the state-of-the-art research on SDGs, focusing on their taxonomies and the use of LLMs for text classification tasks.

\subsection{SDG - Text Classification and Taxonomy Generation}

A review of the scientific literature reveals a variety of NLP techniques applied to document classification within the context of the United Nations' SDGs. Researchers have employed these techniques for diverse purposes, including ontology creation and the development of classification models.

For instance, the OSDG initiative, as described in~\cite{DBLP:journals/corr/abs-2005-14569}, aims to consolidate and advance efforts in categorizing research aligned with the SDGs. The approach integrates features from prior research\footnote{\url{https://github.com/TechNote-ai/osdg/blob/master/OSDG_DATA_SOURCES.md}}, such as ontology elements from the work of Bautista-Puig et al.\footnote{\url{https://figshare.com/articles/dataset/SDG_ontology/11106113/1}}, which developed an ontology of keywords based on UN descriptions and keywords from documents labelled with SDGs. Additionally, it incorporates attributes of machine learning models and keywords extracted from the Seventh Framework Programme for Research and Technological Development\footnote{\url{https://www.fp7-4-sd.eu/tpl/static/FP7\%20and\%20the\%20SDGs\%20-\%20full\%20report\%20(August\%202015).pdf}} (FP7-4-SD project\footnote{\url{https://www.fp7-4-sd.eu/}}). These features are refined and integrated to create a comprehensive OSDG ontology, which is subsequently mapped to topics within the Microsoft Academic Graph\footnote{\url{https://www.microsoft.com/en-us/research/project/microsoft-academic-graph/}} (MAG). This mapping enhances the ontology with rich relationships between terms and underlying concepts.

Additionally, authors in~\cite{joshi2021a} introduced a formal Knowledge Organization System (KOS)~\cite{salatino2024survey} that models the key components of the Global SDG Indicator Framework, which includes 17 Goals, 169 Targets, and 231 unique indicators, along with more than 400 related statistical data series. This ontology is supplemented by a dataset that formalizes and assigns unique identifiers to the SDGs and their indicators, with mappings to terms in the United Nations Bibliographic Information System (UNBIS) and EuroVoc vocabularies. 

In a related study, the authors of~\cite{10.1145/3428502.3428604} developed a classification system that leverages pre-trained state-of-the-art deep learning models, such as the Universal Sentence Encoder, for text classification within the SDG domain. This approach eliminates the need for traditional transfer learning or extensive training procedures. Demonstrated in a legal context, this method effectively classifies UN Resolutions based on their relevance to the SDGs.

Another study performed in~\cite{schmidt_2021_4964606} assessed the accuracy of the Aurora SDG classification model version 5\footnote{\url{https://github.com/Aurora-Network-Global/sdgs_many_berts}} in aligning research papers with the SDGs. This evaluation utilized a baseline set of publications manually selected by researchers, focusing on a specific SDG. Precision and recall metrics were measured and compared not only to its predecessor but also to the Elsevier SDG model\footnote{\url{https://www.elsevier.com/about/sustainability/sdg-research-mapping-initiative}}.

In addition to academic research, several commercial initiatives are addressing SDG text classification. For instance, in collaboration with Springer Nature, a leading publisher of influential journals and a pioneer in open research, Digital Science Consultancy\footnote{\url{https://www.digital-science.com/product/consultancy/}}, a trusted partner providing research workflow solutions and data insights, and Dimensions\footnote{\url{https://www.dimensions.ai/}}, a provider of advanced scientific research databases, have jointly developed a classification system utilizing the Dimensions publications database\footnote{\url{https://www.digital-science.com/}}. This system initially categorized scholarly articles into five SDGs using supervised machine learning with curated training data and later expanded to encompass all 17 SDGs. The training sets are manually refined by experts to address false positives and improve the model's accuracy. Consequently, the system automatically assigns SDG categories to new publications and grants in Dimensions. You can find more information about the Dimensions publications database at \url{https://www.digital-science.com/}.

Finally, SDG Juicer\footnote{\url{https://sdgjuicer.com/}} is a collaborative platform developed by Ovum\footnote{\url{https://ovum.ai/}}, Linkalab\footnote{\url{https://www.linkalab.it/}}, and AB Innovation Consulting\footnote{\url{https://www.abinnovationconsulting.com/}}. This platform, accessible at \url{https://sdgjuicer.com/}, utilizes Artificial Intelligence to extract SDGs from corporate documents, providing tailored recommendations to enhance business impact. Operating as a cloud-based tool, SDG Juicer evaluates the social and environmental impact of companies, aligning their strategies with the UN Agenda 2030. By promoting sustainable and responsible business practices, it contributes to advancing global sustainability goals.

\subsection{Advancements of LLMs for text classification}

Over the past five years, LLMs have significantly advanced NLP by demonstrating an exceptional ability to process large volumes of textual data, enabling efficient information extraction from many different sources, including research papers~\cite{bolanos2024artificial}, patents~\cite{kosonocky2024mining}, medical records~\cite{omiye2024large}, social media posts~\cite{yang2024mentallama}, news articles~\cite{motta2024capturing}, technical documentation~\cite{cascini2004natural}, legal texts~\cite{savelka2023unreasonable},  knowledge graphs~\cite{peng2023knowledge}, tourism-related materials~\cite{cadeddu2024optimizing}, and financial documents~\cite{li2023large}.
Among the areas most transformed by this revolution is text classification, which plays a critical role in structuring and interpreting unstructured data. LLMs have significantly enhanced the accuracy and scalability of classification tasks across multiple domains~\cite{zhang2024pushing}, enabling applications such as sentiment analysis~\cite{miah2024multimodal}, intent recognition~\cite{wang2024beyond}, automated fact-checking~\cite{tang2024minicheck}, and topic-based document organization~\cite{cadeddu2024comparative}.

The integration of transformer architectures into text classification began with the introduction of BERT and GPT in 2018~\cite{devlin-etal-2019-bert,Radford2018ImprovingLU}. This development marked a pivotal moment as indicated by a recent work which indicates that transformer-based models represent approximately 68\% of the top performers in NLP tasks~\cite{10380590}.

The subsequent launch of the HuggingFace platform\footnote{\url{https://huggingface.co/}} provided a user-friendly environment for utilizing several transformer-based models~\cite{jain2022introduction}. 
The widespread adoption of transformer-based NLP approaches significantly increased with the release of ChatGPT in late 2022, which achieved 100 million users within two months\footnote{\url{https://www.theguardian.com/technology/2023/feb/02/chatgpt-100-million-users-open-ai-fastest-growing-app}}. This growing interest has led to more exploration of question-answer-based NLP applications, but the potential for other text classification tasks is still mostly unexplored.

Recent studies have highlighted several state-of-the-art LLMs excelling in unimodal text classification. These investigations examine factors like the size of the training data and the technical nuances among different model architectures, such as encoder-decoder, decoder-only, and encoder-only.~\cite{10.1145/3495162, info13020083}.

Despite the increasing size of LLMs, factors like data volume, privacy concerns, task complexity, and resource availability are crucial when selecting optimal text classification solutions. The high development costs associated with LLMs present a significant barrier, primarily manageable by large tech companies and research institutions with substantial funding\footnote{\url{https://lambdalabs.com/blog/demystifying-gpt-3}}. Democratizing access to this technology depends on improving GPU accessibility and affordability~\cite{10.1145/3649506}. Platforms like Google Colab and Kaggle have made GPUs more accessible, which is essential for running resource-heavy NLP applications~\cite{bisong2019building}.

The increasing advancement of LLMs has underscored the critical need for robust hardware. As the capabilities of language models continue to expand, ensuring widespread access to powerful hardware will be essential for fostering innovation and enabling broader application of these technologies.

Despite the growing interest in utilizing LLMs for text-related tasks, there has not yet been a published comparative evaluation of task adaptation techniques of LLMs specifically focused on the SDG classification task.  Our work aims to fill this gap by providing the first comprehensive analysis of how different LLMs perform on the SDG classification using different strategies. This evaluation will provide useful insights into the strengths and weaknesses of different models, guiding future research and applications in this important area.

\section{Background}
\label{sec:background}

This section describes the OSDG Community Dataset and introduces the eighteen LLMs employed in our experiments.

\subsection{OSDG Community Dataset}
\label{sec:EU_Dataset}

Within the OSDG initiative, the OSDG Community Dataset (OSDG-CD) is a product of the collaborative efforts of over 1,000 volunteers from more than 110 countries. It focuses on aligning text content with the SDGs. These contributors, including researchers, subject-matter experts, and SDG advocates worldwide, have enriched our understanding of SDGs by validating tens of thousands of text excerpts, and assessing their relevance to SDGs based on their background knowledge in labeling exercises. This dataset serves as a valuable resource for gaining insights into SDGs through ontology-based or ML methodologies\footnote{\url{https://zenodo.org/records/5550238}}.

The dataset used in this work, which is updated periodically with new texts, is the version released in October 2023. It contains 42,065 text excerpts and a total of 303,643 assigned labels. Each excerpt is 3 to 6 sentences long, averaging about 90 words, and is sourced from publicly available documents such as reports, policy documents, and publication abstracts. A significant portion of these documents, over 3,000, originates from UN-related sources such as SDG-Pathfinder\footnote{\url{https://sdg.iisd.org/news/oecd-tool-applies-sdg-lens-to-international-organizations-policy-content/}} and SDG Library\footnote{\url{https://www.sdglibrary.ca/}}.

For the purposes of the work described in this article, it is important to highlight the following details:

i) The texts in the dataset were evaluated by volunteers from the OSDG community based on the labels previously assigned to them by SDG-Pathfinder and the SDG Library.

ii) Each volunteer could either accept or reject the label associated with the text assigned to them.

iii) The same text, along with its label, was validated by nine volunteers.

iv) A text, along with its label, was included in the dataset if its label was validated by at least three volunteers, regardless of whether the validation was positive (acceptance of the label) or negative (rejection of the label).

v) For each text included in the dataset, a synthetic indicator of the quality of the label assigned by SDG-Pathfinder and the SDG Library was determined. This indicator, called ``agreement", ranges from 0.0 to 1.0 and is defined as the ratio between the absolute value of the difference between positive and negative evaluations and the total number of evaluations. This indicator measures, in absolute terms, how much the volunteers agreed with each other in evaluating a text, whether positively or negatively. Low values of this indicator indicate that the label of a text was evaluated as positive or negative in very similar or even equal proportions by the volunteers who processed that text, while high values indicate that the label of a text was predominantly evaluated by the volunteers either as positive or as negative. 

Further information on the compilation of the dataset, from which our benchmark dataset used in this work was extracted, can be found on the page of the OSDG-CD\footnote{\url{https://zenodo.org/records/5550238}}.

In detail, the dataset includes the following column descriptions: 
\begin{enumerate}
    \item DOI/ID: Digital Object Identifier of the original document, and the unique text identifier, 
    \item Text: Text excerpt from the document, 
    \item SDG: The SDG class against which the text is validated, 
    \item LN: Number of volunteers rejecting the suggested SDG label, 
    \item LP: Number of volunteers accepting the suggested SDG label, 
    \item Agreem.: Agreement score, calculated as \texttt{abs(LP - LN) / (LP+LN)}.
\end{enumerate}

An example of a few records is illustrated in Table~\ref{fig:sample_of_osdg-community-data-v2023-10-01}.

\sloppy

\begin{table*}[h!]
    \centering
    \resizebox{\textwidth}{!}{
    \begin{tabular}{l|p{8cm}|c|c|c|c}
        \toprule
        \textbf{DOI/ID} & \textbf{Text} & \textbf{SDG} & \textbf{LN} & \textbf{LP} & \textbf{Agreem.} \\ \midrule
        \makecell[l]{DOI: 10.18356/eca72908-en \\ ID: 00028349a7f9b2485ff344ae44ccfd6b} & Labour legislation regulates maximum working hours, industrial safety, minimum wages and benefits for workers and the prevention of child labour, but enforcement, if any, is weak. Therefore, the immigration laws applicable in Western countries cannot be applied to these workers. T. H. including basic public services. & 11 & 2 & 1 & 0.33 \\ \hline
        \makecell[l]{DOI: 10.1787/9789264289062-4-en \\ID: 0004eb64f96e1620cd852603d9cbe4d4} & The average figure also masks large differences across regions in Kazakhstan. The number of annual contacts ranges from 2.0 in Astana to 9.7 in Mangystau, and some parts of the population are likely to have very limited access to primary care. In addition, poor coverage of outpatient prescription medicines limits both the effectiveness (and appeal) of care at PHC level. & 3 & 1 & 8 & 0.78 \\ \hline
        \makecell[l]{DOI: 10.1787/9789264258211-6-en \\ ID: 0006d6e7593776abbdf4a6f985ea6d95} & A region reporting a higher rate will not earn points for this indicator. This means for example that activities that were carried out in 2012 are rewarded financially through the P4P scheme in 2014. The level of payment in the P4P scheme is set deliberately low at the start of the programme to avoid gaming and crowding out intrinsic motivation. & 3 & 2 & 2 & 0 \\ \hline
        \makecell[l]{DOI:10.1787/9789264117563-8-en \\ ID: 000bfb17e9f3a00d4515ab59c5c487e7} & The Israel Oceanographic and Limnological Research station monitors the quantity and quality of water along the coastline of the Mediterranean Sea. The Nature and Parks Authority (NPA) monitors water quality in rivers on behalf of the MoEP. Mekorot and local authorities monitor drinking water quality under the supervision of the Ministry of Health. The Ministry of Health monitors effluent quality prior to its use in the agricultural sector. & 6 & 0 & 3 & 1 \\ \bottomrule
    \end{tabular}
    }
    \caption{Sample of few records from OSDG-community-data-v2023-10-01.}
    \label{fig:sample_of_osdg-community-data-v2023-10-01}
\end{table*}

It is important to note that the original dataset contains texts labeled only for the first 16 SDGs (``SDG 1'' to ``SDG 16''), and does not include texts classified under ``SDG 17'', which pertains to ``Strengthen the means of implementation and revitalize the Global Partnership for Sustainable Development''.  This is because recently there has been a revision of the number of goals and two more goals (SDG-16 and SDG-17) have been added in 2020; consequently, at the time the work described in this article was carried out, the dataset contains data labeled and evaluated by volunteers only for goals 1-16 and not yet for goal 17. Therefore, our analysis focuses solely on the classification of texts related to the first 16 SDGs.

For this study, we required samples with high agreement among annotators and positive feedback (i.e., with a greater number of votes for labels\_positive rather than for labels\_negative) to establish a robust gold standard. However, some texts in the dataset also showed cases where volunteer classifications were assessed negatively with high agreement scores. Therefore, we applied a filter to select only texts where: i) the number of volunteers who gave a positive assessment of the label assigned by the expert evaluator is more than double those who gave a negative assessment. ($\text{labels\_positive} > 2 \times \text{labels\_negative}$), and ii) the agreement on the positive label assessment must be at least 0.75. Following this criterion and ensuring class balance, a dataset with 16 classes, each containing 400 texts, was compiled, totaling 6,400 texts.

Additionally, since the original OSDG Community Dataset does not include texts corresponding to a class where none of the 16 SDGs is identified, which is essential for the classification tasks, we supplemented the original dataset with an additional 400 texts sourced from websites providing online news. We manually verified that these texts did not refer to topics associated with the 16 SDGs, as in the case of this example: ``NYC crime: 3 violent attacks on subways in span of 8 hours in Brooklyn, Queens; No arrests made  WABC-TVView Full Coverage on Google News.''. This verification ensures that, during the execution of the task, the model correctly classifies these texts as ``Other'', which corresponds to the seventeenth category that we have used.

From the resulting dataset of 6,800 text, three balanced subsets were established for the classification task: a training dataset (4,760 texts, 280 per class), a validation dataset (1,020 texts, 60 per class), and a test dataset (1,020 texts, 60 per class). Each set maintained a uniform distribution across the 17 classes, comprising the 16 SDGs and the ``Other'' class also referred to as ``SDG 0''.

\subsection{Used Models}
\label{sec:Used_models}

In our study, we employed a set of 18 language models. The models used include: 

\begin{enumerate} 
    \item bert-base-uncased\footnote{\url{https://huggingface.co/google-bert/bert-base-uncased}}, 
    \item gpt-3.5-turbo-0301\footnote{\url{https://platform.openai.com/docs/models/gpt-3-5-turbo}}, 
    \item gpt-4-0613\footnote{\url{https://platform.openai.com/docs/models/gpt-4-and-gpt-4-turbo}}, 
    \item Llama-2-7b-chat-hf\footnote{\url{https://huggingface.co/meta-llama/Llama-2-7b-chat-hf}}, 
    \item Llama-2-7b-chat-hf\_4-bit\_quantized\footnote{4-bit quantized version of \url{https://huggingface.co/meta-llama/Llama-2-7b-chat-hf}}, 
    \item Llama-2-13b-chat-hf\footnote{\url{https://huggingface.co/meta-llama/Llama-2-13b-chat-hf}}, 
    \item Llama-2-13b-chat-hf\_4-bit\_quantized\footnote{4-bit quantized version of \url{https://huggingface.co/meta-llama/Llama-2-13b-chat-hf}}, 
    \item Llama-2-70b-chat-hf\footnote{\url{https://huggingface.co/meta-llama/Llama-2-70b-chat-hf}}, 
    \item Llama-3\_8B-Instruct\footnote{\url{https://huggingface.co/meta-llama/Meta-Llama-3-8B-Instruct}},
    \item flan\--t5-base\footnote{\url{https://huggingface.co/google/flan-t5-base}}, 
    \item flan-t5-large\footnote{\url{https://huggingface.co/google/flan-t5-large}}, 
    \item flan\--t5-\-xlarge\footnote{\url{https://huggingface.co/google/flan-t5-xl}}, 
    \item flan-t5-xxlarge\footnote{\url{https://huggingface.co/google/flan-t5-xxl}}, 
    \item Mistral-7B-instruct\footnote{\url{https://huggingface.co/mistralai/Mistral-7B-Instruct-v0.1}}, 
    \item Mixtral-8x7B-Instruct-v0.1\footnote{\url{https://huggingface.co/mistralai/Mixtral-8x7B-Instruct-v0.1}},
    \item Ze\-phyr-7b-\-beta\_4bit\footnote{\url{https://huggingface.co/HuggingFaceH4/zephyr-7b-beta}},
    \item Phi-3-mini-4k-instruct\footnote{\url{https://huggingface.co/microsoft/Phi-3-mini-4k-instruct}},
    \item Phi-3-mini-128k-instruct\footnote{\url{https://huggingface.co/microsoft/Phi-3-mini-128k-instruct}}.
 
\end{enumerate}

These models vary significantly in both size, ranging from 220 million to 1.5 trillion parameters, and structure, covering both encoder-decoder and decoder-only models. This diversity enabled a comprehensive evaluation across different scales and configurations.

Due to the considerable size of some models, which precludes loading them into memory even for simple inference tasks, quantized 4-bit versions were utilized in certain instances. These models are accessible through the use of the Quantized LoRA (QLoRA) library~\cite{NEURIPS2023_1feb8787}, which builds upon the LoRA library~\cite{hu2022lora}. Typically, weight parameters in models are stored in a 32-bit format (FP32), where each element of a matrix occupies 32 bits of space. QLoRA significantly reduces memory usage by quantizing the weight parameters down to 4 bits, without greatly compromising the model’s performance compared to its original training.

\subsubsection{BERT}

Introduced by Google researchers in 2018~\cite{devlin-etal-2019-bert}, BERT, an encoder-only Transformer model, quickly set new standards in NLP tasks such as language comprehension, question answering, and named entity recognition. 
 
BERTbase features 12 transformer layers, 12 attention heads, and 110 million parameters, while BERTlarge is equipped with 24 transformer layers, 16 attention heads, and 340 million parameters, delivering superior accuracy~\cite{devlin-etal-2019-bert}. BERT underwent a rigorous four-day Pre-Training period on a large corpus comprising approximately 2.5 billion words from Wikipedia and 800 million words from the Google Books Corpus\footnote{\url{https://www.english-corpora.org/googlebooks/}}. This extensive training allowed BERT to develop proficiency in English and several other languages. Transfer learning techniques optimized the training process, distinguishing the initial extensive Pre-Training from a subsequent fine-tuning phase on task-specific data. The bidirectional training capability of BERT, a feature of its underlying transformer architecture, utilizes a masked language modeling (MLM) technique. By masking 15\% of tokenized words during training, MLM forces the model to predict the masked word based on the context provided by surrounding words, a technique inspired by successful strategies in computer vision.

The model variants, bert-base-uncased and BERT cased, cater to different needs: the uncased model, with its 110 million parameters, does not differentiate between lowercase and uppercase letters (e.g., ``english'' vs. ``English''), making it suitable for general text classification tasks, whereas the cased model retains original text casing and accent markers, vital for applications such as Named Entity Recognition and Part-of-Speech tagging.

\subsubsection{GPT-3.5 and GPT-4.0}

The GPT-3.5-turbo-0301 model\footnote{\url{https://platform.openai.com/docs/models/gpt-3-5-turbo}}, a decoder-only framework with 175 billion parameters, marks a refined iteration of the GPT-3.5 series, enhanced through Reinforcement Learning with Human Feedback (RLHF). This model supports up to 4096 tokens, covering both input and output elements. GPT-3, the foundation of GPT-3.5, was trained on a diverse array of datasets, notably including 60\% of its foundational training data from a curated version of the Common Crawl dataset\footnote{\url{https://commoncrawl.org/}}. Other significant data sources include WebText2\footnote{WebText2 is not publicly available; however, an open-source equivalent, OpenWebText2, was created using similar data sources and is available at \url{https://openwebtext2.readthedocs.io/en/latest/}}, Books1 and Books2, which together account for 16\% of the training data\footnote{Books1 and Books2 details can be found at \url{https://our-hometown.com/overview-of-data-used-to-train-language-models/}}. GPT-3.5 Turbo was specifically tailored for improved performance in dialogues, question-and-answer sessions, and interactive commands, including code generation and execution.

The GPT-4-0613 model\footnote{\url{https://platform.openai.com/docs/models/gpt-4-and-gpt-4-turbo}} is an advanced multimodal model, capable of processing and generating text from both image and text inputs. It demonstrated human-level proficiency across various benchmarks, significantly outperforming GPT-3.5 in a simulated bar exam by scoring in the top 10\%~\cite{openai2024gpt4}. Developed over six months with extensive iterative alignment and adversarial testing, GPT-4 is better aligned with user instructions and safety protocols. Despite its improvements, the differences between GPT-3.5 and GPT-4 in everyday tasks may seem subtle, but GPT-4's reliability, creativity, and nuanced handling of complex tasks set it apart. It is important to highlight that GPT-4 is built using an architecture based on a ``mixture of experts'', consisting of eight models, each with 220 billion parameters. This structure allows for specialized handling of tasks, managed by a meta-model that optimizes performance and integrates outputs effectively.

\subsubsection{LLaMa-2 and LLaMa-3}

LLaMa-2, developed by Meta, is a decoder-only Transformer encompassing a range of pre-trained and subsequently fine-tuned generative text models with sizes varying from 7 billion to 70 billion parameters~\cite{touvron2023llama}. The LLaMa-2-Chat variants, specifically tailored for dialogue applications, have demonstrated superior performance over several open-source chat models in benchmarks conducted by their developers. 
 
Utilizing an auto-regressive framework and an optimized transformer architecture, LLaMa-2 models are engineered for efficient language processing. Their Fine-Tuning process includes Supervised Fine-Tuning (SFT) and Reinforcement Learning with Human Feedback (RLHF), aligning them closely with human preferences in terms of helpfulness and safety. The entire LLaMa-2 series was pre-trained on a massive corpus of 2 trillion tokens derived from publicly accessible sources\footnote{\url{https://llama.meta.com/llama2/}}, and the Fine-Tuning phase incorporated publicly available instruction datasets\footnote{\url{https://www.ibm.com/topics/instruction-tuning}}\footnote{\url{https://huggingface.co/datasets/HuggingFaceH4/instruction-dataset}} and over a million newly created human-annotated examples. Notably, neither the Pre-Training nor the Fine-Tuning datasets incorporate user data from Meta. The larger models employ Grouped-Query Attention (GQA)~\cite{ainslie-etal-2023-gqa}, enhancing their scalability in inference tasks. While these models are primarily designed for assistant-like chat functions, the pre-trained variants offer versatility for various natural language generation tasks. Proper usage of these models for chat-oriented applications requires adherence to specific formatting guidelines, including the use of INST and SYS tags, beginning-of-sentence (BOS) and end-of-sentence (EOS) tokens, and appropriate spacing and line breaks.

Recently, Meta has unveiled two next-generation models\footnote{\url{https://ai.meta.com/blog/meta-llama-3/}}: LLaMa-3-8B, an 8-billion parameter system with a knowledge cutoff in March 2023, and LLaMa-3-70B, a 70-billion parameter system with its knowledge base updated until December 2023. The latest iteration now supports a context window of 8,192 tokens, up from 4,096, enabling the processing of longer prompts. Trained on 15 trillion tokens, which is seven times more data than LLaMa 2, these models boast a vastly expanded vocabulary that enhances the efficient encoding of both input and output text, promising improved multilingual performance and overall capabilities. To maximize LLaMa 3’s potential in chat and dialogue applications, Meta introduced an innovative instruction fine-tuning method that blends supervised fine-tuning (SFT) with rejection sampling, proximal policy optimization (PPO), and direct preference optimization (DPO). Meta also confirmed that the training data was curated multiple times to mitigate biases before its application in SFT and related tuning processes.

\subsubsection{Flan-T5}

The Flan-T5 models~\cite{chung2022scaling} represent a fine-tuned evolution of the original T5 models, which were encoder-decoder systems initially trained on a mix of supervised and unsupervised tasks;

in this type of model, a large corpus of text data is employed to predict missing words in an input text using a fill-in-the-blank objective~\cite{JMLR:v21:20-074}. This iterative process continues until the model learns to generate text that closely resembles the input data. Once trained, FLAN-T5 can be applied to a range of NLP tasks, including text generation, language translation, sentiment analysis, and text classification.
This 
Fine-Tuning process included tasks from the GLUE\footnote{GLUE - \url{https://gluebenchmark.com/}} and SuperGLUE\footnote{SuperGLUE - \url{https://super.gluebenchmark.com/}} benchmarks as well as token manipulation where 15\% of tokens were randomly replaced with sentinel tokens. 

The models come in various sizes: Flan-T5-base has 220 million parameters, Flan-T5-large has 770 million, Flan-T5-xl boasts 3 billion, and Flan-T5-xxl includes 11 billion parameters, all with a context capacity of 512 tokens.

\subsubsection{Mistral, Mixtral and Zephyr}

Mistral AI, a French company specializing in AI products, was established in April 2023 by former employees of Meta Platforms and Google DeepMind\footnote{\url{https://en.wikipedia.org/wiki/Mistral_AI}}. The company launched its first LLM, the Mistral-7B-v0.1, which features a decoder-based architecture and 7.3 billion parameters~\cite{jiang2023mistral}. This model introduces several architectural innovations: first, the Sliding Window Attention mechanism, which can handle a context length of 8,000 tokens and has a fixed cache size, theoretically allowing it to attend to up to 128,000 tokens. Second, it incorporates Grouped Query Attention (GQA), which enhances inference speed while reducing cache size. Third, it utilizes a Byte-fallback Byte Pair Encoding (BPE) tokenizer to ensure consistent character recognition without resorting to out-of-vocabulary tokens. The Mistral-7B-Instruct-v0.1 variant, specifically fine-tuned for instructional purposes, is optimized for chat-based applications. This Fine-Tuning was conducted using publicly available instructional datasets from HuggingFace and did not include any proprietary data. Notably, this model surpasses the performance of the LLaMa 2 13B model across all benchmarks introduced in~\cite{jiang2023mistral}.

Following the success of its initial model, Mistral AI introduced a second LLM, the Mixtral-8x7B~\cite{jiang2024mixtral}. This model is a high-quality sparse mixture of experts (SMoE) with open weights, licensed under Apache 2.0. Mixtral stands out for its superior performance compared to the LLaMa 2 70B model in most benchmarks, offering six times faster inference with only 7 billion parameters. It is recognized as the most powerful model with open weights under a permissive license, excelling in cost-performance trade-offs and matching or outperforming GPT-3.5 in standard benchmarks~\cite{jiang2024mixtral}. The Mixtral-8x7B shares several features with Mistral AI's first model, including Sliding Window Attention with a training context length of 8,000 tokens and Grouped Query Attention for enhanced inference speed. The Byte-fallback BPE tokenizer also remains a feature, ensuring accurate character recognition. The Mixtral-8x7B-Instruct-v0.1, like its predecessor, is an instruction fine-tuned model geared toward chat-based inference applications.

Hugging Face leveraged the success of Mistral 7B to develop Zephyr-7B Alpha~\cite{tunstall2023zephyr}, highlighting that a fine-tuned Mistral 7B could surpass much larger chat models and even compete with GPT-4 in specific tasks. This release marked just the beginning, as Zephyr-7B Beta followed soon after.
Zephyr 7B enhances its responsiveness and alignment with human instructions by harnessing the capabilities of larger models through knowledge distillation. This technique involves training smaller models on the complex patterns learned by larger ones, thereby reducing training requirements without compromising language modeling performance.
In this process, the ``student" model learns to perform tasks that were previously out of reach by absorbing the detailed patterns of the ``teacher" model. Rather than focusing solely on final predictions, the student is trained to replicate the teacher’s output probabilities or features, allowing it to capture the subtle decision-making processes of the teacher. This approach often leads to improved performance compared to training with only ground truth data. 
In developing Zephyr-7B, two primary datasets were employed to train and refine the model, each targeting different facets of dialogue generation. The first, the UltraChat Dataset\footnote{\url{https://huggingface.co/datasets/HuggingFaceH4/ultrachat_200k}}, is derived from GPT-3.5-turbo-generated dialogues and comprises 1.47 million multi-turn conversations spanning 30 topics and 20 types of text material. The second, the UltraFeedback Dataset\footnote{\url{https://huggingface.co/datasets/HuggingFaceH4/ultrafeedback_binarized}}, includes 64,000 prompts, each accompanied by four responses, that were evaluated by GPT-4 based on criteria such as instruction-following, honesty, and helpfulness. Together, these datasets are essential for training Zephyr-7B to understand and generate dialogue that is both human-like and aligned with high standards of instruction adherence, honesty, and helpfulness.

\begin{table*}[t!]
\centering
\footnotesize
\caption{Models employed across the various learning approaches.\label{tab:experiments_explained}}
\begin{tabular}{lccccccccccccccccccc}
model         & \Rot{bert-base-uncased} & \Rot{gpt-3.5-turbo-0301} & \Rot{gpt-4-0613} & \Rot{Llama-2-7b-chat-hf} & \Rot{Llama-2-7b-chat-hf\_4-b\_quanz} & \Rot{Llama-2-13b-chat-hf} & \Rot{Llama-2-13b-chat-hf\_4-b\_quanz} & \Rot{Llama-2-70b-chat-hf} & \Rot{Llama-3-8b-chat} & \Rot{flan-t5-base} & \Rot{flan-t5-large} & \Rot{flan-t5-xlarge} & \Rot{flan-t5-xxlarge} & \Rot{Mistral-7B-instruct} & \Rot{Mixtral-8x7B-Instr.-v0.1} & \Rot{Zephyr-7b-beta\_4bit} & \Rot{Phi-3-mini-4k-instruct} & \Rot{Phi-3-mini-128k-instruct} \\
\midrule
ZSL           &                   & x                  & x          & x                  & x                                    & x                   & x                                     & x & x                  & x            & x             & x              & x               & x                   & x                          & x  & x & x                 \\ \hline
FSL (3, 5, 7) &                   & x                  & x          & x                  & x                                    & x                   & x                                     & x & x                  &              &               &                &                 & x                   & x                          & x &  x & x                  \\ \hline
FT            & x                 & x                  & x          & x                  & x                                    & x                   & x                                     &   & x                  & x            & x             & x              & x                & x                   & x                          & x &  x & x                 \\
\bottomrule
\end{tabular}
\end{table*}

\begin{table*}[t!]
    \centering
    \scriptsize
    \caption{Results of the experiments when employing ZSL. Values are in percentages.}\label{tab:ZS}
        \begin{tabular}{l|rrrr}
        \toprule
            \textbf{MODEL NAME} & \textbf{Pre} & \textbf{Rec} & \textbf{Acc} & \textbf{F1} \\
            \midrule        

            \textbf{gpt-3.5-turbo-0301} & \textbf{83.1} & \textbf{81.8} & \textbf{81.8} & \textbf{81.1} \\
            gpt-4-0613 & 82.1 & 80.7 & 80.7 & 80.3 \\
            \hdashline
            Llama-2-7b-chat-hf\_4bit & 60.9 & 44.8 & 44.8 & 41.8 \\
            Llama-2-7b-chat-hf & 63.3 & 54.7 & 54.7 & 53.8 \\
            Llama-2-13b-chat-hf\_4bit & 70.7 & 65.7 & 65.7 & 65.5 \\
            Llama-2-13b-chat-hf & 69.4 & 68.4 & 68.4 & 67.0 \\
            Llama-2-70b-chat-hf & 76.5 & 68.5 & 68.5 & 69.6 \\
            Llama-3\_8B-Instruct & 75.4 & 65.6 & 65.6 & 66.0 \\
            \hdashline
            flan-t5-base (220M) & 68.0 & 41.4 & 41.4 & 43.4 \\
            flan-t5-large (800M) & 71.3 & 30.7 & 30.7 & 35.5 \\
            flan-t5-xlarge (3B) & 80.8 & 52.3 & 52.3 & 57.6 \\
            flan-t5-xxlarge (11B) & 80.3 & 74.4 & 74.4 & 74.8 \\
            \hdashline
            Mistral\_7B\_instruct & 64.3 & 55.6 & 55.6 & 51.6 \\
            Mixtral-8x7B-Instruct & 75.2 & 67.1 & 67.1 & 68.7 \\
            Zephyr-7b-beta\_4bit & 72.0 & 67.4 & 67.4 & 67.6 \\
            \hdashline
            Phi-3-mini-4k-instruct & 69.7 & 59.2 & 59.2 & 59.8 \\
            Phi-3-mini-128k-instruct & 62.4 & 48.4 & 48.4 & 48.9 \\      
            \bottomrule
        \end{tabular}
\end{table*}

\subsubsection{Phi-3}

Phi-3~\cite{abdin2024phi3}, a cutting-edge series of decoder-only small language models developed by Microsoft in April 2024, represents the pinnacle of affordability and capability in AI. These models excel in various benchmarks, including language processing, reasoning, coding, and mathematics, surpassing their larger counterparts~\cite {abdin2024phi3}. 
The Phi-3-mini, with its 3.8 billion parameters, is a standout model available in 4K and 128K token context-length options. It notably supports up to a 128K token context window, the first in its category to do so without significant quality degradation.
These models are instruction-tuned to intuitively follow diverse commands mirroring everyday communication, ensuring they are user-friendly.
The Phi-3 model range includes Phi-3-mini, Phi-3-small with 7 billion parameters, and Phi-3-medium with 14 billion parameters. The development of these models adhered to Microsoft's Responsible AI Standard, which encompasses accountability, transparency, fairness, reliability, safety, privacy, security, and inclusiveness. Extensive safety assessments, red-teaming, sensitive use reviews, and security compliance were integral to their development, aligning with Microsoft's stringent best practices.
Phi-3 models benefit from training on high-quality data and enhancements from extensive post-training safety measures, including reinforcement learning from human feedback (RLHF), automated testing, and manual red-teaming across numerous harm categories.
Their compact size makes Phi-3 models particularly suited for environments with limited resources, such as on-device and offline scenarios, or situations demanding quick response times and cost-effectiveness, especially for simpler tasks.
Phi-3-mini excels in compute-limited environments and, due to its modest computational demands, offers a cost-effective solution with superior latency. Its extended context window facilitates comprehensive analysis and reasoning over substantial textual content like documents, web pages, and code.
The Phi-3-mini-4k-instruct model, featuring 4-bit OmniQuant quantization, provides a 4K context length (dependent on available memory). It is compatible for download on all iPhones and iPads with at least 6GB of RAM and on Intel or Apple Silicon Macs, enhancing device capabilities with powerful AI chatbot functionalities.

\section{Experimental Evaluation}
\label{sec:experimentalevaluation}

The main objective of this study was to evaluate the performance of various types of LLMs on the benchmark we created out of the OSDG Community Dataset.

Specifically, we systematically analyzed 18 different language models to classify text excerpts from the OSDG Community dataset according to the 17 SDGs mentioned earlier.  
To achieve this, we employed two different task adaptation techniques: 
\begin{itemize}
    \item \textbf{in-Context Learning}: Zero-Shot Learning (ZSL), Few-Shot Learning (FSL), and
    \item \textbf{Fine-Tuning (FT)}.
\end{itemize}

In ZSL, models directly process the test set using a basic prompt without examples. For each model, various prompt types were tested to determine the most effective. 

In FSL, the prompts include a number, \( S \), of sample texts from the training set. We evaluated model's performance using three distinct \( S \) values of samples: 3, 5, and 7. In order to identify the most suitable sample of excerpts to include within the prompt, we explored three different strategies: 
\begin{enumerate}
    \item \textit{RANDOM} - \( S \) samples were randomly chosen from the training set for each test set element. 
    \item \textit{Semantic Similarity (SEM. SIMILARITY)} - samples from the training set were ranked in descending order of their semantic similarity to the text associated with the underlying test set element, selecting the top \( S \) samples (without the constraint that examples must come from different classes). We compute the pair-wise similarity between sentences using cosine similarity.
    \item \textit{Semantic Similarity with Different Classes (SS WITH DC)} - similarly ranked by semantic similarity, but with a constraint that samples from the training set must come from \( S \) different classes, introducing more class variety into the model's evaluation process during FSL. In contrast, method ii) often results in selecting training texts from the same class as the test text, limiting diversity.
\end{enumerate}

\begin{table*}[t!]
    \centering
    \caption{FSL with 3 texts; texts have been extracted from the training dataset randomly (``RANDOM'' columns), by semantic similarity with the processed text (``SEM. SIMILARITY'' columns) and by semantic similarity with the processed text with the constraint of belonging all to different classes (``SS WITH DC'' columns).}
    \label{tab:FS_3}

\addtolength{\tabcolsep}{-0.3em}
        \begin{tabular}{l|cccc|cccc|cccc}
        \toprule
        & \multicolumn{4}{c|}{\textbf{RANDOM}} & \multicolumn{4}{c|}{\textbf{SEM. SIMILARITY}} & \multicolumn{4}{c}{\textbf{SS WITH DC}} \\
       
        \midrule

        \textbf{MODEL NAME} & \textbf{Pre} & \textbf{Rec} & \textbf{Acc} & \textbf{F1} & \textbf{Pre} & \textbf{Rec} & \textbf{Acc} & \textbf{F1} & \textbf{Pre} & \textbf{Rec} & \textbf{Acc} & \textbf{F1} \\
        \midrule        

        gpt-3.5-turbo-0301 & 80.5 & 80.2 & 80.2 & 79.0 & 84.7 & 83.8 & 83.8 & 83.4 & 82.9 & 81.6 & 81.6 & 81.3 \\
        gpt-4-0613 & 81.8 & 80.1 & 80.1 & 79.7 & 86.1 & 85.6 & 85.6 & 85.4 & 84.2 & 83.2 & 83.2 & 83.1 \\
        \hdashline
        Llama-2-7b-chat-hf\_4bit & 65.0 & 45.4 & 45.4 & 47.2 & 76.6 & 73.5 & 73.5 & 72.8 & 53.6 & 45.2 & 45.2 & 44.1 \\
        Llama-2-7b-chat-hf & 61.4 & 48.1 & 48.1 & 48.5 & 78.5 & 76.2 & 76.2 & 75.9 & 63.7 & 58.9 & 58.9 & 58.1 \\
        Llama-2-13b-chat-hf\_4bit & 70.2 & 69.9 & 69.9 & 68.5 & 84.8 & 84.2 & 84.2 & 83.9 & 74.1 & 71.4 & 71.4 & 70.8 \\
        Llama-2-13b-chat-hf & 67.7 & 69.3 & 69.3 & 67.3 & 83.5 & 83.2 & 83.2 & 82.7 & 73.9 & 73.3 & 73.3 & 72.5 \\
        Llama-2-70b-chat-hf & 73.9 & 71.1 & 71.1 & 70.2 & 82.1 & 80.9 & 80.9 & 80.2 & 79.7 & 78.0 & 78.0 & 77.8 \\
        \textbf{Llama-3\_8B-Instruct} & 68.0 & 55.4 & 55.4 & 56.9 & \textbf{86.0} & \textbf{86.0} & \textbf{86.0} & \textbf{86.0} & 83.7 & 82.4 & 82.4 & 82.5 \\

        \hdashline
        Mistral\_7B\_instruct & 66.5 & 58.2 & 58.2 & 54.9 & 84.4 & 82.9 & 82.9 & 82.5 & 76.5 & 72.1 & 72.1 & 72.1 \\
        Mixtral-8x7B-Instruct & 81.0 & 79.8 & 79.8 & 79.2 & 86.2 & 85.5 & 85.5 & 85.2 & 84.0 & 82.5 & 82.5 & 82.4 \\
        Zephyr-7b-beta\_4bit & 72.1 & 68.1 & 68.1 & 68.5 & 72.6 & 68.5 & 68.5 & 69.3 & 70.4 & 67.0 & 67.0 & 67.3 \\ 
        \hdashline
        Phi-3-mini-4k-instruct & 75.0 & 72.6 & 72.6 & 72.7 & 84.3 & 84.0 & 84.0 & 84.0 & 72.3 & 69.5 & 69.5 & 69.9 \\
        Phi-3-mini-128k-instruct & 73.0 & 70.8 & 70.8 & 70.9 & 77.0 & 75.4 & 75.4 & 75.5 & 67.0 & 63.2 & 63.2 & 63.9 \\
        
        \bottomrule
        \end{tabular}
\end{table*}

\begin{table*}[t!]
    \centering
    \caption{FSL with 5 texts; texts have been extracted from the training dataset randomly (``RANDOM'' columns), by semantic similarity with the processed text (``SEM. SIMILARITY'' columns) and by semantic similarity with the processed text with the constraint of belonging all to different classes (``SS WITH DC'' columns).}
    \label{tab:FS_5}
\addtolength{\tabcolsep}{-0.3em}
        \begin{tabular}{l|cccc|cccc|cccc}
        \toprule
        & \multicolumn{4}{c|}{\textbf{RANDOM}} & \multicolumn{4}{c|}{\textbf{SEM. SIMILARITY}} & \multicolumn{4}{c}{\textbf{SS WITH DC}} \\
        
        \midrule

        \textbf{MODEL NAME} & \textbf{Pre} & \textbf{Rec} & \textbf{Acc} & \textbf{F1} & \textbf{Pre} & \textbf{Rec} & \textbf{Acc} & \textbf{F1} & \textbf{Pre} & \textbf{Rec} & \textbf{Acc} & \textbf{F1} \\
        \midrule    

        gpt-3.5-turbo-0301 & 81.8 & 80.7 & 80.7 & 79.7 & 84.7 & 84.2 & 84.2 & 83.9 & 83.1 & 82.2 & 82.2 & 81.9 \\
        \textbf{gpt-4-0613} & 83.1 & 81.6 & 81.6 & 81.2 & \textbf{87.5} & \textbf{87.1} & \textbf{87.1} & \textbf{86.9} & 84.7 & 83.7 & 83.7 & 83.7 \\
        \hdashline
        Llama-2-7b-chat-hf\_4bit & 62.2 & 52.3 & 52.3 & 51.6 & 78.3 & 76.1 & 76.1 & 75.9 & 59.1 & 53.6 & 53.6 & 52.3 \\
        Llama-2-7b-chat-hf & 64.2 & 58.7 & 58.7 & 58.9 & 80.2 & 78.9 & 78.9 & 78.6 & 69.7 & 67.1 & 67.1 & 66.4 \\
        Llama-2-13b-chat-hf\_4bit & 72.3 & 70.5 & 70.5 & 69.1 & 86.1 & 85.9 & 85.9 & 85.3 & 74.4 & 73.4 & 73.4 & 73.0 \\
        Llama-2-13b-chat-hf & 68.2 & 68.9 & 68.9 & 66.9 & 85.9 & 85.2 & 85.2 & 84.8 & 75.3 & 74.7 & 74.7 & 73.7 \\
        Llama-2-70b-chat-hf & 74.5 & 70.7 & 70.7 & 70.2 & 85.0 & 83.7 & 83.7 & 83.3 & 79.3 & 77.7 & 77.7 & 77.3 \\
        Llama-3\_8B-Instruct & 68.9 & 62.8 & 62.8 & 62.2 & 86.5 & 85.9 & 85.9 & 85.9 & 83.2 & 82.3 & 82.3 & 82.3 \\
        
        \hdashline
        Mistral\_7B\_instruct & 66.3 & 60.1 & 60.1 & 56.6 & 86.2 & 84.6 & 84.6 & 84.2 & 76.7 & 71.0 & 71.0 & 70.9 \\
        Mixtral-8x7B-Instruct & 79.3 & 78.8 & 78.8 & 78.0 & 87.1 & 86.5 & 86.5 & 86.4 & 81.6 & 79.9 & 79.9 & 79.8 \\
        Zephyr-7b-beta\_4bit & 74.5 & 70.6 & 70.6 & 70.8 & 76.7 & 72.9 & 72.9 & 73.9 & 74.2 & 71.9 & 71.9 & 72.1 \\
        \hdashline
        Phi-3-mini-4k-instruct & 76.9 & 69.8 & 69.8 & 71.0 & 85.5 & 85.1 & 85.1 & 85.1 & 73.3 & 69.5 & 69.5 & 70.3 \\
        Phi-3-mini-128k-instruct & 73.4 & 71.1 & 71.1 & 71.1 & 81.5 & 81.2 & 81.2 & 81.1 & 69.4 & 66.2 & 66.2 & 66.9 \\
        
        \bottomrule
               
        \end{tabular}
\end{table*}

\begin{table*}[t!]
    \centering
    \caption{FSL with 7 texts; texts have been extracted from the training dataset randomly (``RANDOM'' columns), by semantic similarity with the processed text (``SEM. SIMILARITY'' columns) and by semantic similarity with the processed text with the constraint of belonging all to different classes (``SS WITH DC'' columns).}
    \label{tab:FS_7}
\addtolength{\tabcolsep}{-0.3em}
        \begin{tabular}{l|cccc|cccc|cccc}
        \toprule
        & \multicolumn{4}{c|}{\textbf{RANDOM}} & \multicolumn{4}{c|}{\textbf{SEM. SIMILARITY}} & \multicolumn{4}{c}{\textbf{SS WITH DC}} \\
        
        \midrule

        \textbf{MODEL NAME} & \textbf{Pre} & \textbf{Rec} & \textbf{Acc} & \textbf{F1} & \textbf{Pre} & \textbf{Rec} & \textbf{Acc} & \textbf{F1} & \textbf{Pre} & \textbf{Rec} & \textbf{Acc} & \textbf{F1} \\
        \midrule        

        gpt-3.5-turbo-0301 & 82.0 & 81.3 & 81.3 & 80.4 & 86.3 & 85.9 & 85.9 & 85.7 & 82.9 & 82.1 & 82.1 & 81.8 \\
        gpt-4-0613 & 83.3 & 81.7 & 81.7 & 81.4 & 87.9 & 87.5 & 87.5 & 87.4 & 84.7 & 83.9 & 83.9 & 83.8 \\
        \hdashline
        Llama-2-7b-chat-hf\_4bit & 69.1 & 48.8 & 48.8 & 51.5 & 83.5 & 83.0 & 83.0 & 82.8 & 66.2 & 63.8 & 63.8 & 63.1 \\
        Llama-2-7b-chat-hf & 65.6 & 52.2 & 52.2 & 53.5 & 83.5 & 82.4 & 82.4 & 82.1 & 71.0 & 68.8 & 68.8 & 68.3 \\
        Llama-2-13b-chat-hf\_4bit & 70.0 & 59.8 & 59.8 & 61.2 & 85.5 & 84.6 & 84.6 & 84.8 & 70.5 & 69.4 & 69.4 & 69.1 \\
        Llama-2-13b-chat-hf & 70.5 & 69.9 & 69.9 & 68.1 & 87.0 & 86.5 & 86.5 & 86.2 & 73.4 & 73.3 & 73.3 & 72.1 \\
        Llama-2-70b-chat-hf & 75.1 & 72.9 & 72.9 & 71.9 & 85.7 & 84.9 & 84.9 & 84.1 & 78.7 & 77.7 & 77.7 & 77.1 \\
        Llama-3\_8B-Instruct & 68.3 & 55.7 & 55.7 & 55.8 & 86.4 & 85.6 & 85.6 & 85.7 & 84.8 & 83.8 & 83.8 & 83.9 \\   
        \hdashline
        Mistral\_7B\_instruct & 67.1 & 63.9 & 63.9 & 62.3 & 86.9 & 85.6 & 85.6 & 85.3 & 77.2 & 71.7 & 71.7 & 71.6 \\
        \textbf{Mixtral-8x7B-Instruct} & 78.5 & 78.5 & 78.0 & 78.0 & \textbf{88.8} & \textbf{88.2} & \textbf{88.2} & \textbf{88.1} & 83.4 & 81.8 & 81.8 & 81.7 \\
        Zephyr-7b-beta\_4bit & 69.0 & 61.2 & 61.2 & 62.7 & 76.8 & 72.8 & 72.8 & 73.7 & 71.4 & 68.6 & 68.6 & 68.8 \\
        \hdashline
        Phi-3-mini-4k-instruct & 76.0 & 69.1 & 69.1 & 70.5 & 85.1 & 84.6 & 84.6 & 84.5 & 72.9 & 71.1 & 71.1 & 71.2 \\
        Phi-3-mini-128k-instruct & 72.8 & 59.8 & 59.8 & 62.3 & 83.7 & 83.3 & 83.3 & 83.3 & 69.8 & 64.8 & 64.8 & 65.9 \\
        
        \bottomrule
        
        \end{tabular}
\end{table*}

\begin{table}[t!]
    \centering
    \caption{Results of the experiments when employing Fine-Tuning.}
    \label{tab:FT}

        \begin{tabular}{l|cccc}

            \toprule
            \textbf{MODEL NAME} & \textbf{Pre} & \textbf{Rec} & \textbf{Acc} & \textbf{F1} \\
            \midrule            

            gpt-3.5-turbo-0301 & 91.7 & 91.6 & 91.6 & 91.4  \\
            \hdashline
            Llama-2-7b-chat-hf\_4bit & 85.4 & 84.8 & 84.8 & 84.8 \\
            Llama-2-7b-chat-hf & 86.8 & 85.3 & 85.3 & 85.5 \\
            Llama-2-13b-chat-hf\_4bit & 86.5 & 85.6 & 85.6 & 85.7 \\
            \textbf{Llama-2-13b-chat-hf} & \textbf{92.5} & \textbf{92.4} & \textbf{92.4} & \textbf{92.4} \\
            Llama-3\_8B-Instruct & 87.6 & 86.8 & 86.8 & 86.7 \\
            \hdashline
            flan-t5-base (220M) & 90.5 & 90.6 & 90.6 & 90.5 \\
            flan-t5-large (800M) & 90.7 & 90.6 & 90.6 & 90.5 \\
            flan-t5-xlarge (3B) & 90.9 & 90.8 & 90.8 & 90.7 \\
            flan-t5-xxlarge (11B) & 91.1 & 90.9 & 90.9 & 90.9 \\
            \hdashline
            bert\_base\_uncased & 87.1 & 87.3 & 87.3 & 87.2 \\
            \hdashline
            Mistral\_7B\_instruct & 88.0 & 88.2 & 88.2 & 88.1 \\
            Mixtral-8x7B-Instruct & 85.7 & 85.3 & 85.3 & 85.2 \\
            Zephyr-7b-beta\_4bit & 86.9 & 86.4 & 86.4 & 86.4 \\
            \hdashline
            Phi-3-mini-4k-instruct & 70.7 & 62.7 & 62.7 & 63.5 \\
            Phi-3-mini-128k-instruct & 66.8 & 56.8 & 56.8 & 56.7 \\
            
            \bottomrule
        \end{tabular}
\end{table}

\begin{table*}[t!]
    \centering
    \caption{Summary of our results based on the F1 Score for ZSL, FT, and ZSL focused on the \textit{SEM. SIMILARITY} method.}
    \label{tab:Summary}

        \begin{tabular}{lccccc}
        \toprule
            \small\textbf{MODEL NAME} & \small\textbf{ZS} & \small\textbf{FS-SS-3} & \small\textbf{FS-SS-5} & \small\textbf{FS-SS-7} & \small\textbf{FT} \\
            \midrule

                bert\_base\_uncased & - & - & - & - & 87.2 \\
                \hdashline
                \textbf{gpt-3.5-turbo-0301} & \textbf{81.1} & 83.4 & 83.9 & 85.7 & 91.4 \\
                \textbf{gpt-4-0613} & 80.3 & 85.4 & \textbf{86.9} & 87.4 & - \\
                \hdashline
                Llama-2-7b-chat-hf\_4bit & 41.8 & 72.8 & 75.9 & 82.8 & 84.8 \\
                Llama-2-7b-chat-hf & 53.8 & 75.9 & 78.6 & 82.1 & 85.5 \\
                Llama-2-13b-chat-hf\_4bit & 65.5 & 83.9 & 85.3 & 84.8 & 85.7 \\
                \textbf{Llama-2-13b-chat-hf} & 67.0 & 82.7 & 84.8 & 86.2 & \textbf{92.4} \\
                Llama-2-70b-chat-hf & 69.6 & 80.2 & 83.3 & 84.1 & - \\
                \textbf{Llama-3\_8B-Instruct} & 66.0 & \textbf{86.0} & 85.9 & 85.7 & 86.7 \\
                \hdashline
                flan-t5-base (220M) & 43.4 & - & - & - & 90.5 \\
                flan-t5-large (800M) & 35.5 & - & - & - & 90.5 \\
                flan-t5-xlarge (3B) & 57.6 & - & - & - & 90.7 \\
                flan-t5-xxlarge (11B) & 74.8 & - & - & - & 90.9 \\
                \hdashline
                Mistral\_7B\_instruct & 51.6 & 82.5 & 84.2 & 85.3 & 88.1 \\
                \textbf{Mixtral-8x7B-Instruct} & 68.7 & 85.2 & 86.4 & \textbf{88.1} & 85.2 \\
                Zephyr-7b-beta\_4bit & 67.6 & 69.3 & 73.9 & 73.7 & 86.4 \\
                \hdashline
                Phi-3-mini-4k-instruct & 59.8 & 84.0 & 85.1 & 84.5 & 63.5 \\
                Phi-3-mini-128k-instruct & 48.9 & 75.5 & 81.1 & 83.3 & 56.7 \\
                
                \bottomrule
        \end{tabular}
\end{table*}

Finally, in FT, the pre-trained models were subjected to an additional training stage to incorporate a task-specific layer onto the existing architecture. This FT process utilized 
the training dataset and validation dataset created from the OSDG-CD and described in Section \ref{sec:EU_Dataset}.
The performance of the FT models was subsequently evaluated on the 
same test dataset created from the OSDG-CD and described in Section \ref{sec:EU_Dataset}, also used for the ZSL and FSL experiments.

All ZSL and FSL prompts are publicly available at the following URL: \url{https://tinyurl.com/sdg-paper-prompts}.

Given all the possible combinations of models and learning approaches, we omitted certain experiments that were either unfeasible due to the context window limitations of some language models or too demanding computationally. Table~\ref{tab:experiments_explained} summarises the employed models across the three learning approaches.

Specifically, we conducted ZSL experiments on all models except BERT, as it requires Fine-Tuning prior to task application.
For FSL we performed experiments with all models except: i) BERT as it requires pre-task Fine-Tuning and ii) Flan-T5 as its maximum sequence length of tokens is too small (only 512 tokens) and therefore it truncates the prompts when they are too long, as it often happens in the case of FSL experiments~\cite{JMLR:v21:20-074}.

Due to resource limitations, we conducted ZSL experiments on all models except LLaMa-2-70b-chat-hf, as it requires a cluster of 8 A40 Large GPUs with 48 GB of vRAM each, for a total of 384 GB of vRAM.

All models have been fine-tuned using prompts that presented all training dataset texts with their labels, reflecting its Pre-Training methodology, except for BERT and Flan-T5 models which were fine-tuned by simply providing the training texts with their labels, without using prompts, according to models' characteristics.

The FT experiments adhered to typical parameter settings found in the literature for the models in this study, such as the number of epochs, batch size, learning rate, dropout rate, and others. For BERT, 5 epochs were used with a learning rate of $2 \cdot 10^{-5}$ and a batch size of 8. In contrast, for LLaMa-2 and Flan-T5, 1 epoch was employed along with a learning rate of $2 \cdot 10^{-4}$, a batch size of 4, a LoRA attention dimension of 64, an Alpha parameter for LoRA scaling of 16, a Dropout probability for LoRA layers of 0.1, a dtype of float16 for 4-bit base models, and a quantization type of nf4. For GPT-3.5, LLaMa-3 and Mixtral, 3 epochs were employed. Mistral was fine-tuned employing 10 epochs, with a learning rate of $10^{-5}$ and a batch size of 8. Zephir was fine-tuned employing 2 epochs, with a learning rate of $3 \cdot 10^{-5}$, a batch size of 32, and an Alpha parameter for LoRA scaling of 8. Phi-3 models were fine-tuned employing 1 epoch, with a learning rate of $5 \cdot 10^{-6}$, a batch size of 4, an Alpha parameter for LoRA scaling of 32, and a Dropout probability for LoRA layers of 0.05.

\section{Results}
\label{sec:results}

\color{black}
In this section, we detail the results of our experiments. We start with an analysis of the model performance in the Zero-Shot Learning (ZSL) setting, followed by the outcomes using Few-Shot Learning (FSL), and conclude with the results from FT. Finally, we summarize and discuss the most significant findings.

The models were evaluated based on Precision, Recall, Accuracy, and F1-score. These metrics were calculated using the macro-average approach, which is particularly suitable in this context as it treats all classes as equally important. 

Table \ref{tab:ZS} report the performance of the models in the ZSL setting. In this category, the GPT family stands out, with GPT-3.5 achieving 81.1\%, surprisingly, this is marginally higher than GPT-4’s 80.3\%.  
This is likely because GPT models have been trained on a much larger corpus and possess a significantly greater number of parameters than other models, enhancing their information-processing capabilities in the ZSL context. They also likely had access to a substantial amount of text in the SDG domain during Pre-Training, enabling greater competence regarding the selected context. Notably, the Flan-T5-xxl model with 11 billion parameters also performs quite well (74.8\%), surpassing well-known alternatives such as LLaMa-2 and Mistral. 
This finding aligns with the result from~\cite{wei2022finetuned}, where Flan-T5 is demonstrated to be competitive with GPT-3 in various ZSL tasks\footnote{\url{https://exemplary.ai/blog/flan-t5}}.

\color{black}

Tables \ref{tab:FS_3}, \ref{tab:FS_5}, and \ref{tab:FS_7} report the performance of the models using FSL with three, five, and seven exemplary texts. 
The examples for FSL were drawn from the training dataset, using the three methods described in Section \ref{sec:experimentalevaluation}. Notably, the \textit{RANDOM} method yielded results comparable to those of ZSL experiments, suggesting that a random sample may yield no advantage for this task. 
The \textit{SEM. SIMILARITY} method generally produced significantly better outcomes. This difference is attributed to the fact that randomly selected examples may not provide the models with sufficient context for task resolution. In contrast, semantically informative examples enhance the models' predictive capabilities by deepening their understanding of the text-label relationships. Notably, the \textit{SEM. SIMILARITY} method showed an average F1-score improvement of 19.3\% (3 samples), 21.5\% (5 samples), and 20.9\% (7 samples). The \textit{SS with DC} method, where texts must come from different classes, yielded lower gains of 10.0\% (3 texts), 11.9\% (5 texts), and 12.5\% (7 texts). The inferior performance were likely due to occasionally excluding semantically closer texts in favour of diversity. 
The results suggest that prioritizing semantic proximity is more beneficial for this task.

Different models excel with varying numbers of examples. Specifically, LLaMa-3\_8B-Instruct performs best with only 3 examples (86.0\%), GPT-4.0 excels with 5 examples (86.9\%), and Mixtral-8x7B-Instruct is superior with up to 7 examples (88.1\%). These findings suggest that FSL is a complex scenario where the optimal model may depend on several factors. They also indicate that the combination of innovative mechanisms implemented in recent models, such as LLaMa-3 and Mixtral, including mobile attention, supervised FT, rejection sampling, proximal policy optimization, and direct preference optimization, can be highly beneficial. This can lead relatively smaller models to outperform models with trillions of parameters, such as GPT-4.0, in certain contexts. 
However, the GPT models exhibited more consistent performance across all three example selection approaches compared to alternative models. This consistency suggests greater robustness to changes in selection strategies. In contrast, other models showed significant variations in performance in response to these strategies.

Table \ref{tab:FT} presents the results from the FT experiments.
All relevant models exhibit a significant improvement in performance, showing an average increase in the F1 score of 25.7\% points compared to the ZSL. Notably, the top-performing model is LLaMa-2-13b-chat-hf, which achieved 92.4\%, followed closely by GPT-3.5-turbo at 91.4\% and Flan-T5-XXL at 90.7\%. It is also worth noting that all fine-tuned T5 models displayed strong performance. Of particular note is the Flan-T5-base, a model with 220 million parameters, which achieved an impressive 90.5\%. BERT also performed relatively well, obtaining an F1 of 87.2\%

These results corroborate previous research indicating that smaller models, such as T5, can achieve excellent outcomes, sometimes comparable to much larger models, when fine-tuned on high-quality data~\cite{edwards2024language}. Similarly, the performance of the BERT model aligns with numerous studies in the literature~\cite{edwards2024language, yu2023open}, demonstrating that despite its smaller size and simpler architecture, it performs exceptionally well in certain downstream tasks, even surpassing models with over 7 billion parameters.

Table \ref{tab:Summary} summarizes the results, showcasing the top-performing models across various tests utilizing ZSL, FSL and FT. FSL  refers to experiments conducted using the \textit{SEM. SIMILARITY} approach, which demonstrated superior performance compared to other alternatives. The results provide several interesting insights. First, GPT-3.5 and GPT-4 significantly outperformed all other alternatives when no reference examples were included in the prompt. This is likely due to their size, as they presumably have a robust internal representation of the SDGs.

Second, the advantage of a very large language model like GPT-4 diminishes in an FSL setting. GPT-4 shows better performance only when the prompt includes five examples. In other cases, smaller models such as LLaMa-3\_8B-Instruct and Mixtral-8x7B-Instruct perform better.

Third, the fine-tuned LLaMa-2 13B achieves the best results among all models, including the fine-tuned GPT-3.5, which is somewhat surprising. One possible explanation is its higher number of parameters (13B) compared to recent alternatives like Mistral (7B), Flan-t5-xxlarge (11B), and LLaMa-3 (8B).

Finally, the quantized versions performed quite well given their smaller size. Notably, the fine-tuned versions of LLaMa-2 13B 4-bit (85.7\%) and Zephyr 7B 4-bit (86.4\%) achieved highly competitive results, demonstrating their value in low-resource settings.

\section{Conclusions}
\label{sec:conclusions}

In this study, we evaluated various proprietary and open-source LLMs for a single-label multi-class text classification task targeting the SDGs. Our aim was to investigate whether smaller models, optimized through prompt engineering techniques, could rival the performance of larger, more resource-intensive models such as OpenAI's GPT. 

Our key contributions include the creation of a specialized benchmark dataset for SDG classification tasks, and a thorough comparison of eighteen LLMs under different task adaptation techniques: ZSL, FSL, and FT. We also investigate three techniques for selecting FSL examples: random selection (``RANDOM''), semantic similarity (``SEM. SIMILARITY''), and semantic similarity across different classes (``SS WITH DC'').

The results of our experiments offer several insights:

\begin{itemize}
\item Smaller models (e.g., LLaMa 2, Mistral, Phi), when properly optimized with high-quality FSL, can achieve performance levels comparable or superior to larger models such as GPT-3.5 and GPT 4. 

\item Different sample selection techniques can result in significantly different performances. In this case, a pure semantic similarity metric (\textit{SEM. SIMILARITY}) clearly outperformed both random selection (\textit{RANDOM}) and a more diverse version of semantic similarity (\textit{SS WITH DC}). Notably, the performance achieved with a random selection was comparable to ZSL, underscoring how poor sample selection can lead to no improvement.

\item FT proved to be a powerful method for enhancing model performance across all tested LLMs. The fine-tuned models consistently outperformed their non-fine-tuned counterparts, underscoring the value of domain-specific training in achieving high classification accuracy.

\item Our analysis of quantized versions of the models revealed that reducing precision in neural network weights does not significantly degrade performance. This finding is encouraging for deploying LLMs in resource-constrained environments, where computational efficiency is crucial.

\end{itemize}

Future research will aim to extend this work by exploring additional optimization techniques and evaluating model performance on a broader range of classification tasks in different domains.

\section*{Acknowledgment}
We acknowledge financial support under the National Recovery and Resilience Plan (NRRP), Mission 4 Component 2 Investment 1.5 - Call for tender No.3277 published on December 30, 2021 by the Italian Ministry of University and Research (MUR) funded by the European Union – NextGenerationEU. Project Code ECS0000038 – Project Title eINS Ecosystem of Innovation for Next Generation Sardinia – CUP F53C22000430001- Grant Assignment Decree No. 1056 adopted on June 23, 2022 by the Italian Ministry of University and Research (MUR).

\begin{IEEEbiography}[{\includegraphics[width=1in,height=1.25in,clip,keepaspectratio]{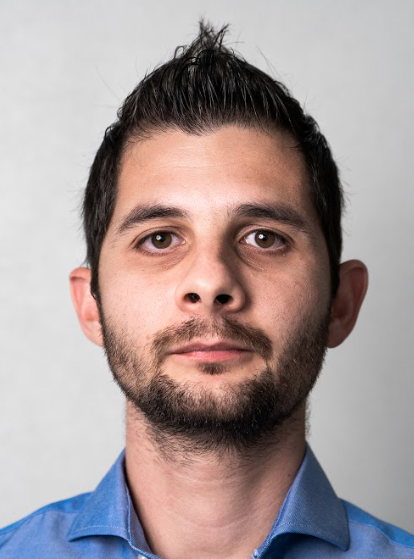}}]{Andrea Cadeddu} currently works as a data scientist at Linkalab, a private Computational Laboratory on Complex Systems. He has a degree in economics and business management and a specialization in data science. He is currently focused on the study and application of NLP models for text classification and topic detection.
\end{IEEEbiography}

\begin{IEEEbiography}[{\includegraphics[width=1in,height=1.25in,clip,keepaspectratio]{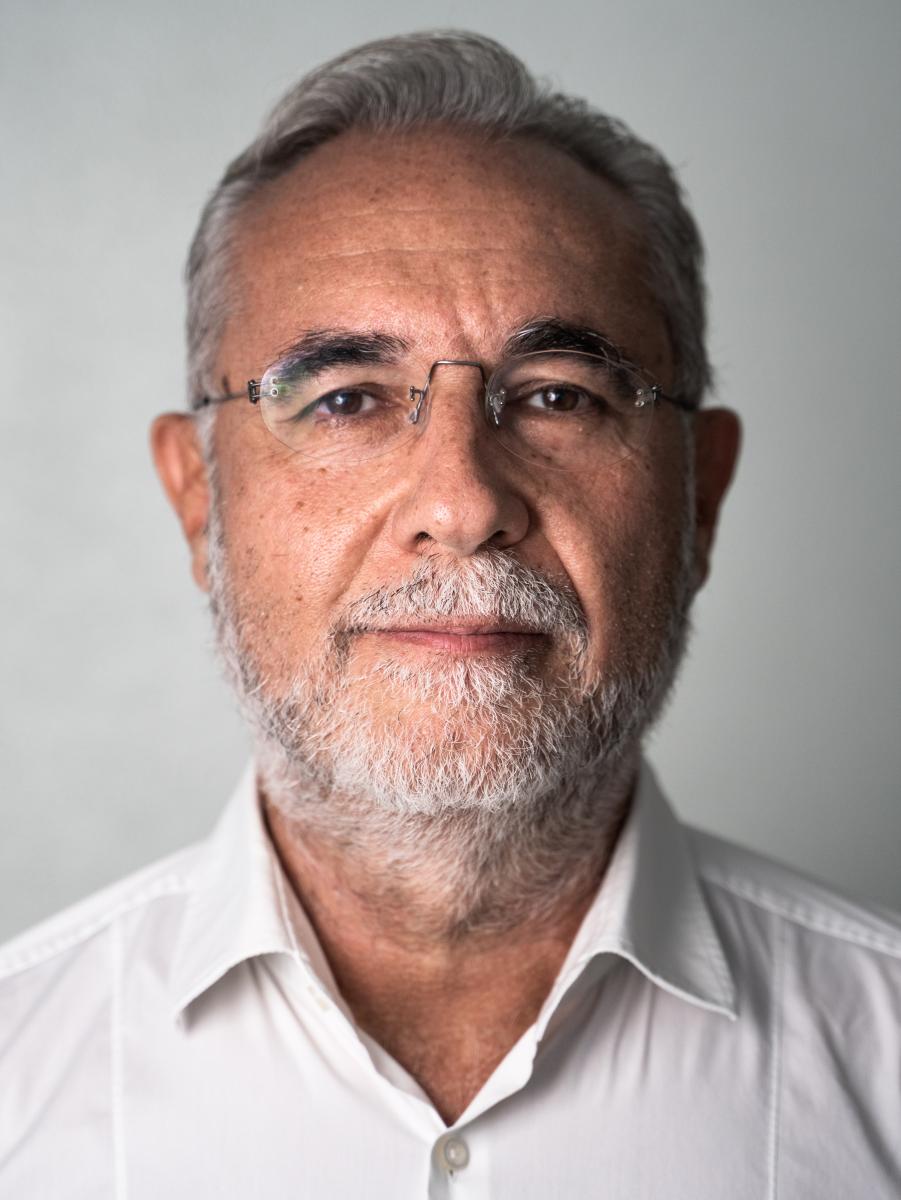}}]{Alessandro Chessa} is Adjunct Professor at Luiss in Data Science \& Artificial Intelligence and at NABA in Machine Creativity. He is currently CEO of Linkalab, a Complex Systems Computational Laboratory, and Scientific Advisor of Eni Datalab. He has been a Research Associate at Boston University working on Econophysics. His scientific interests range from applying Quantum Mechanics to the World Wide Web to the study of the Social Graphs of the new communities on the Internet, and the Data-Driven Journalism. Recently, he has been studying the impact of Artificial Intelligence on human creativity.
\end{IEEEbiography}

\begin{IEEEbiography}[{\includegraphics[width=1in,height=1.25in,clip,keepaspectratio]{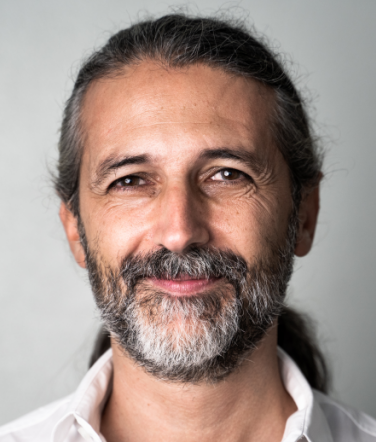}}]{Vincenzo De Leo} is a Ph.D. in High-Energy Physics and is a current Ph.D. student in Computer Science at the Department of Mathematics and Computer Science, University of Cagliari, Italy. After a specialization in Bioinformatics Technologies Applied to Personalized Medicine obtained in 2006 at the CRS4 research laboratory, in Sardinia, he started a position at Linkalab, a private Computational Laboratory on Complex Systems, as a Data Scientist where he actually performs his research studies, in collaboration with the Department of Mathematics and Computer Science of the University of Cagliari, in the field of knowledge injection in Large Language Models. Starting from 2016, he has also been a tenured high-school teacher initially in Computer Science and then, from 2020, in Mathematics and Physics.
\end{IEEEbiography}

\begin{IEEEbiography}[{\includegraphics[width=1in,height=1.25in,clip,keepaspectratio]{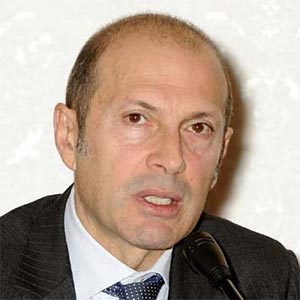}}]{Gianni Fenu} is a full professor of Computer Science at the Department of Mathematics and Computer Science – University of Cagliari (ITALY). He received the ``laurea degree'' in Engineering at the University of Cagliari (ITALY) in 1985, and he joined the University of Cagliari (ITALY) in 1988. He teaches courses in Computer Networks and Information Systems for First Level Degree in Computer Science students, and Network Architecture for Specialized Degree Course. From 2008 to 2015 he was the Coordinator of the Course of Studies of Computer Science. He has been the Director of Bioinformatics and Innovation and Informatics Services Masters of the University of Cagliari. He is the Scientific Responsible of Smart Cities Project of E-learning Ileartv MIUR-UE (2014 -2017, 10 ME, 6 partners). He is currently involved in two Regional Project Natura 2000 (L.R. 7/2007) and in the European Research M-Commerce and Development.
\end{IEEEbiography}

\begin{IEEEbiography}[{\includegraphics[width=1in,height=1.25in,clip,keepaspectratio]{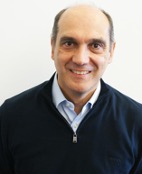}}]{Prof. Enrico Motta} is Professor in Knowledge Technologies and Former Director (2000 -2007) of the Knowledge Media Institute (KMi) at the Open University in UK. Prof. Motta has a Laurea in Computer Science from the University of Pisa in Italy and a PhD in Artificial Intelligence from the Open University. His research spans a variety of aspects at the intersection of large-scale data integration and modelling, semantic and language technologies, intelligent systems, and human-computer interaction. Over the years, Prof. Motta has led KMi's contribution to numerous high-profile projects, receiving over £10.4M in external funding since 2000 from a variety of institutional funding bodies and commercial organizations.
\end{IEEEbiography}

\begin{IEEEbiography}[{\includegraphics[width=1in,height=1.25in,clip,keepaspectratio]{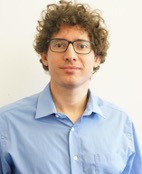}}]{Dr. Francesco Osborne} is a Research Fellow at the Knowledge Media Institute of the Open University in UK, where he leads the Scholarly Data Mining team. He is also Assistant Professor at the University of Milano Bicocca. His research covers Artificial Intelligence, Information Extraction, Knowledge Graphs, Science of Science, and Semantic Web. He authored more than ninety peer-reviewed publications in top journals and conferences of these fields. He collaborates with major publishers, universities, and companies in the space of innovation for producing a variety of innovative services for supporting researchers, editors, and research polities makers. He released many well-adopted resources such as the Computer Science Ontology and the Artificial Intelligence Knowledge Graph. 
\end{IEEEbiography}

\begin{IEEEbiography}[{\includegraphics[width=1in,height=1.25in,clip,keepaspectratio]{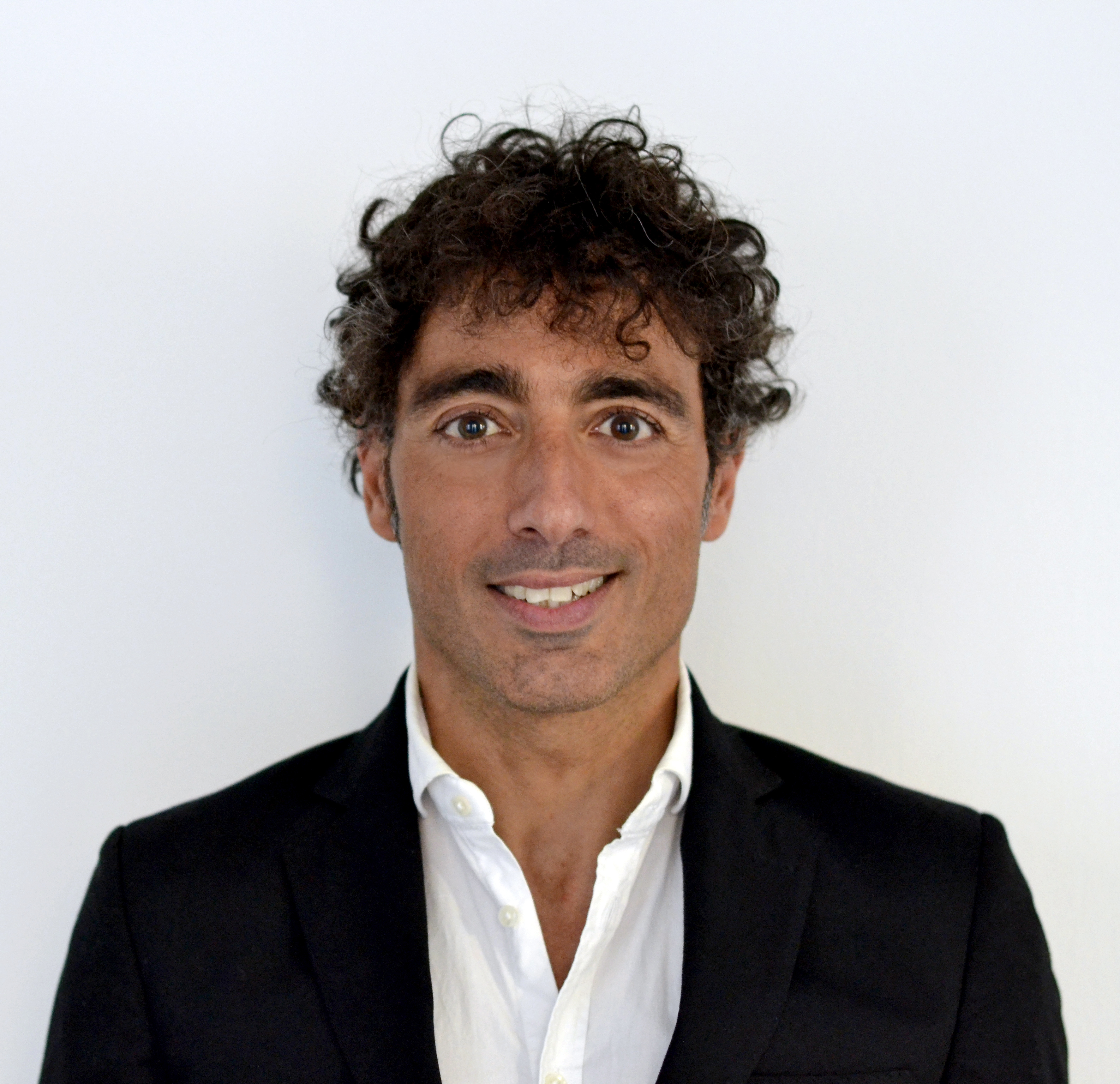}}]{Prof. Diego Reforgiato Recupero} has been a Full Professor at the Department of Mathematics and Computer Science of the University of Cagliari, Italy, since February 2022. He received a Ph.D. degree in computer science from the University of Naples Federico II, Italy, in 2004. From 2005 to 2008 he was a Postdoctoral Researcher at the University of Maryland College Park, USA. 
He won different awards in his career (such as the Marie Curie International Reintegration Grant, Marie Curie Innovative Training Network, Best Researcher Award from the University of Catania, Computer World Horizon Award, Telecom Working Capital, Startup Weekend, Best Paper Award). He co-founded 7 companies within the ICT sector and is actively involved in European projects and research (with one of his companies he won more than 40 FP7 and H2020 projects). His current research interests include sentiment analysis, semantic web, natural language processing, human-robot interaction, financial technology, and smart grid. He is the author of more than 270 conference and journal papers in these research fields, with more than 4800 citations.
\end{IEEEbiography}

\begin{IEEEbiography}[{\includegraphics[width=1in,height=1.25in,clip,keepaspectratio]{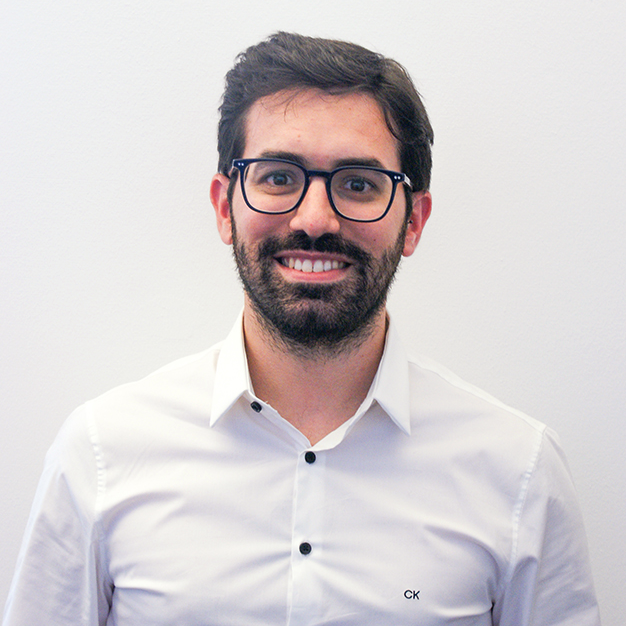}}]{Dr. Angelo Salatino} is a Research Fellow at The Open University who develops AI solutions to analyse scholarly data. He collaborates with Springer Nature to improve the classification and management of research content, and his main research areas include detecting emerging research trends and creating semantic technologies to organise scholarly knowledge. He has created several systems used by publishers and research organisations, including the Computer Science Ontology (CSO), currently the largest taxonomy of research topics in Computer Science; the CSO Classifier, which annotates research documents; and Augur, a tool for detecting emerging research topics. He is particularly interested in exploring how AI can be used to improve understanding of scientific content and support researchers and practitioners in their daily activities. 
\end{IEEEbiography}

\begin{IEEEbiography}[{\includegraphics[width=1in,height=1.25in,clip,keepaspectratio]{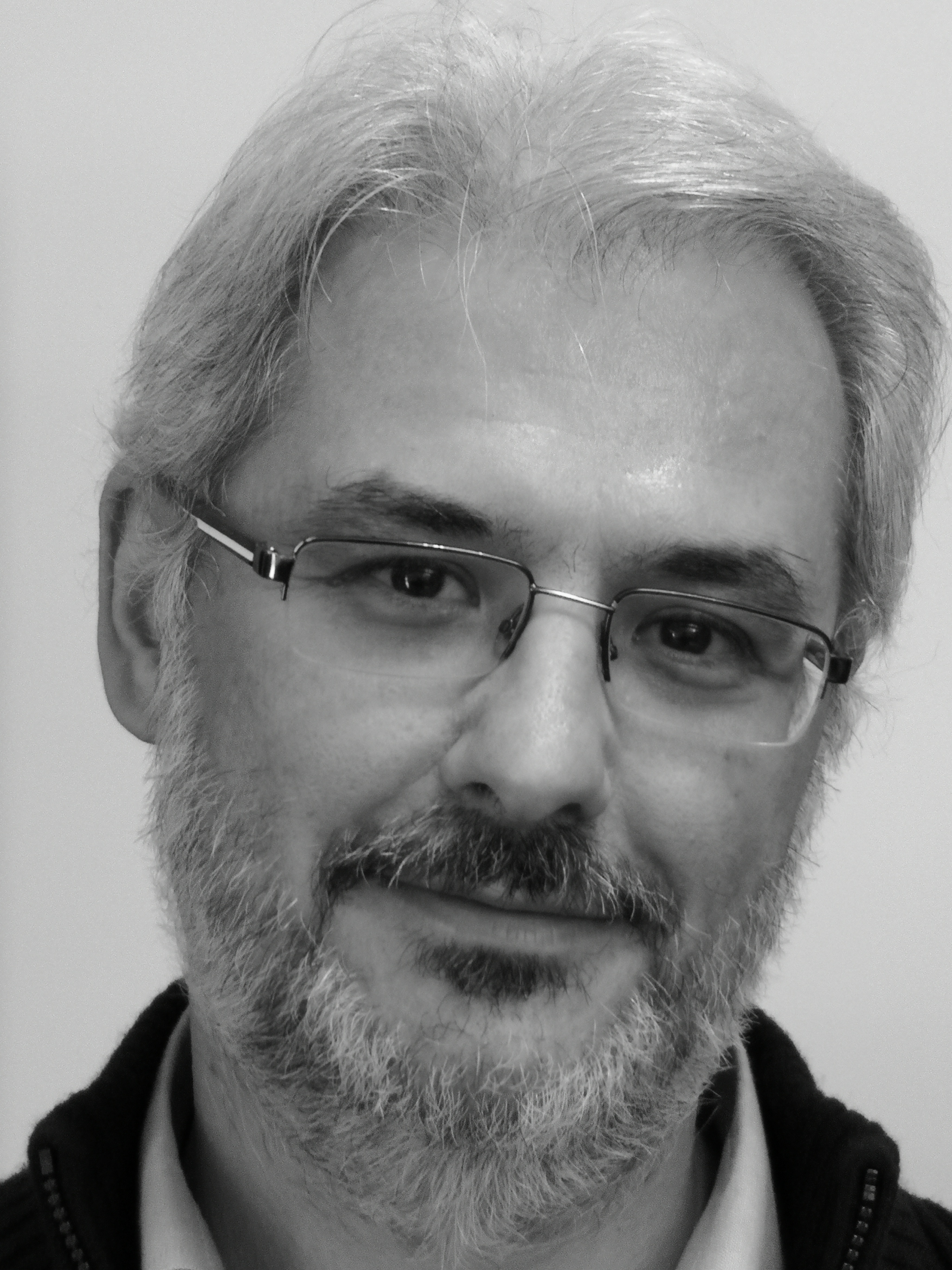}}]{Luca Secchi} received the ''laurea degree`` in Electronic Engineering from the University of Cagliari, Italy and a Ph.D. in Computer Science at the Department of Mathematics and Computer Science, University of Cagliari, Italy. His research interests are in the areas of Knowledge Graphs, Natural Language Processing, Big Data, and the Semantic Web. He has more than 20 years of experience as a professional in the IT field both in the public and private sector. He is one of the partners and the Chief Research and Development Officer of Linkalab, a private Computational Laboratory on Complex Systems.
\end{IEEEbiography}

\EOD

\end{document}